\def\year{2018}\relax
\documentclass[letterpaper]{article} 
\usepackage{aaai18}  
\usepackage{times}  
\usepackage{helvet}  
\usepackage{courier}  
\usepackage{url}  
\usepackage{graphicx}  
\usepackage{multirow}
\usepackage{amsfonts}
\usepackage{CJKutf8}
\usepackage{epsfig}
\usepackage{color}

\begin{document} 
%
\title{Multi-channel Encoder for Neural Machine Translation}
\author{Hao Xiong, Zhongjun He, Xiaoguang Hu and Hua Wu\\
Baidu Inc. No. 10, Shangdi 10th Street, Beijing, 100085, China\\
{\{xionghao05, hezhongjun, huxiaoguang, wu\_hua\} }@baidu.com\\
}
\maketitle
\begin{abstract}

Attention-based Encoder-Decoder has the effective architecture for neural machine translation (NMT), which typically relies on recurrent neural networks (RNN) to build the blocks that will be lately called by attentive reader during the decoding process. This design of encoder yields relatively uniform composition on source sentence, despite the gating mechanism employed in encoding RNN. On the other hand, we often hope the decoder to take pieces of source sentence at varying levels suiting its own linguistic structure: for example, we may want to take the entity name in its raw form while taking an idiom as a perfectly composed unit. Motivated by this demand, we propose Multi-channel Encoder (MCE), which enhances encoding components with different levels of composition. More specifically, in addition to the hidden state of encoding RNN, MCE takes 1) the original word embedding for raw encoding with no composition, and 2) a particular design of external memory in Neural Turing Machine (NTM) for more complex composition, while all three encoding strategies are properly blended during decoding. Empirical study on Chinese-English translation shows that our model can improve by 6.52 BLEU points upon a strong open source NMT system: DL4MT\footnote{\url{https://github.com/nyu-dl/dl4mt-tutorial}}.
On the WMT14 English-French task, our single shallow system achieves BLEU=38.8, comparable with the state-of-the-art deep models.

\end{abstract}
\section{Introduction}
Attention-based neural machine translation has arguably the most effective architecture for neural machine translation (NMT), outperforming conventional statistical machine translation (SMT) systems on many language pairs \cite{uedin-nmt:2017}.  The superiority of attention-based model over canonical encoder-decoder model~\cite{sutskever2014sequence} lies in the fact that it can dynamically retrieve relevant pieces of the source (much isomorphic to alignment in SMT) through a relatively simple matching function.  In other words, attention-based model benefits from a richer representation of source sentence with its flexibility representing local structure.

In a typical attention-based NMT system, a bidirectional recurrent neural networks (biRNN) \cite{schuster1997bidirectional} is used to encode the source, yielding a sequence of vectors from the RNN,  which can be roughly interpreted as context-aware embedding of the words in the source sentence.  With this design, the encoder learns relatively uniform composition of the sentence, despite the RNN in the encoder are already equipped with some advanced gating mechanism, such as long short term memory (LSTM) \cite{hochreiter1997long} network and gated recurrent unit (GRU) \cite{cho2014learning}.  For translation,  it is common that we hope the decoder to take pieces of source sentence at varying composition levels suiting its own linguistic structure.  This need can be illustrated through the following two examples
\begin{itemize}
\item we may want to take the entity name in the source in its raw form while taking an idiom as a densely composed unit;
\item when we translating the substantive, we may again want to know the surrounding words of the noun and then determine its singular or plural forms. 
\end{itemize}

Motivated by this demand, we propose Multi-channel Encoder (MCE), which takes encoding components with different levels of composition. More specifically, in addition to the hidden state of encoding RNN, MCE takes the original word embedding for raw encoding with no composition, and a particular design of external memory in NTM \cite{graves2014neural} for more complex composition, in a way analogous to visual channels with different frequency.  All three encoding strategies are properly blended during decoding controlled by parameters can be learned in an end-to-end fashion. More specifically, we design a gate that can automatically tunes the weights of different encoding channels. 
\begin{figure*}[t]
\centering
\includegraphics[width=7in]{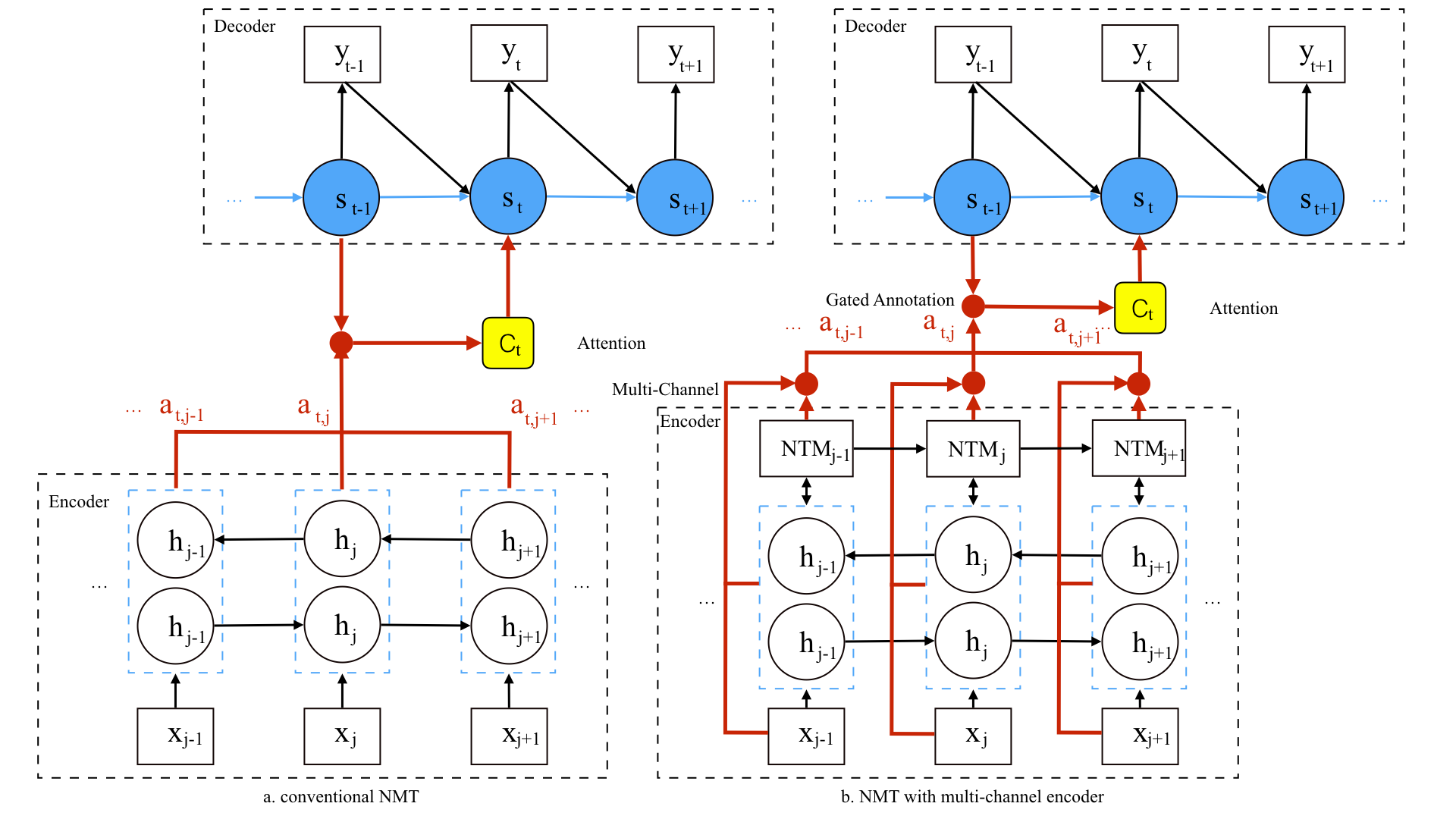}

\caption{Left is the conventional attention-based NMT which consists of encoder, decoder and an attention mechanism. Right is our novel NMT model with multiple channel encoder, which consists of hidden state from biRNN, external memory in the NTM and embeddings directly from the input sequence. A gated annotation is designed to automatically learn the weights for different encoding components.}
\label{fig:main}
\end{figure*}

In this paper, we also empirically investigate MCE on different translation tasks. We first test our models on the NIST Chinese-English machine translation tasks, achieving an average improvement by 6.52 BLEU over the strong DL4MT system. Furthermore, experiments on the WMT2014 English-French machine translation task show that our single shallow model obtains a BLEU score of 38.8, which is comparable to the state-of-the-art models.

The reminder of this paper is organized as follows: in the next section, we will briefly introduce the basics of conventional attention-based NMT. After that, we will present MCE in more details. After that, we will report extensive experimental results and conclude the paper in the last section.

\section{Attention-based Neural Machine Translation}
\label{section2}
In recent years, researchers have proposed excellent works to improve the performance of NMT. Most state-of-the-art NMT systems are based on the attention-based encoder-decoder architecture.
Figure {\ref{fig:main}} (a) illustrates the attention-based encoder-decoder architecture, which consists of three parts: an encoder, a decoder and an attention mechanism that building relationships between encoder and decoder. 

In more details, the first step of NMT systems is to convert each atomic symbol into a corresponding continuous vector, named word embedding. This step is done for each source word independently of the other words and results in a source sequence of word embeddings.
Above the embedding layer, a biRNN is designed to learn the representation of the whole sequence after running on some time steps. Between the encoder and decoder, an attention mechanism is employed to fuse all time steps of the input sequence and draw the attention for current time step in the decoder. During the generation of the target word, the controller will consider the suggestion from last generated word, current hidden state, and the context computed by the attention mechanism to decide the next word.   

Formally, given a source sequence $x=(x_1, ..., x_t)$  and previous translated words $({y_1}, ..., {y_{j-1}})$, the probability of next word $y_j$ is calculated as:
\begin{equation}
p(y_{j}|s_j,y_{j-1},c_j) = \mathrm{softmax}(t_jW_o)
\label{eq:py_j}
\end{equation}
and 
\begin{equation}
t_j = \mathrm{tanh}(s_jW_{t1}+e_{y_{j-1}}W_{t2}+c_jW_{t3})
\label{eq:t_j}
\end{equation}
where $W_{t1}$, $W_{t2}$, $W_{t3}$, $W_o$ are the trained model parameters.
$e_{x_t}$, $e_{y_{j-1}}$ are the embedding representation of $x_t$ and $y_{j-1}$ respectively, which is usually initialized with a one-hot embedding vector. $s_j$ is the hidden state in the  decoder at time step $j$, which is computed as:
\begin{equation}
s_j = g(s_{j-1}, e_{y_{j-1}}, c_j)
\label{eq:s_j}
\end{equation}

Here $g$ is a nonlinear transform function, which can be implemented as LSTM or GRU, and $c_j$ is a distinct context vector at time 
step $j$, which can be obtained by an attention mechanism. Normally,  $c_j$ is calculated as a weighted sum of the input annotations $\mathrm{h}_i$:
\begin{equation}
c_i = \sum_{i}^{Tx} \alpha_{ij}\mathrm{h}_i
\label{eq:c_i}
\end{equation}

where $\mathrm{h}_i = [\overrightarrow{\mathrm{h}_{i}^T}, \overleftarrow{\mathrm{h}_{i}^T}]^T$ is the annotation of $x_i$ from a biRNN and $Tx$ is the length of the source sequence. 
The normalized weight $\alpha_{ij}$ for $\mathrm{h}_i$ is calculated as:
\begin{eqnarray}
\label{eq:alpha_ij}
\alpha_{ij} &=& \frac{\mathrm{exp}(e_{ij})}{\sum_{k=1}^{Tx}\mathrm{exp}(e_{kj})}	\\
\label{eq:e_ij}
e_{ij} &=& V^{T}_{a}\mathrm{tanh}(U_{a}s^{'}_{j}+W_{a}h_{i})
\end{eqnarray}
where $V_a$, $U_a$ and $W_a$ are the trainable parameters. All of the parameters in the
NMT model are optimized to maximize the following
conditional log-likelihood of the M sentence
aligned bilingual samples:
\begin{equation}
\ell(\theta) = \frac{1}{M}\sum^M_{m=1}{\sum^{Ty}_{j=1}{\mathrm{log}p(y_{j}|s_j,y_{j-1},c_j)}}
\label{eq:loss}
\end{equation}

Here, all the biases are omitted for simplify.

\section{Multi-channel Encoder}
\label{sec:3}
As an important part of the attention-based NMT models, the RNN encodes the representation of the source sequence which is lately used by the attention mechanism. 
Nevertheless, as we have mentioned in the first section, it is difficult for the conventional RNN to encode sequence with different levels of composition which is necessary during the translation. Thus, we propose to use multiple-channel to enhance the encoder and the attention mechanism. Figure \ref{fig:main} (b) illustrates the overall architecture of our model, where an external memory is designed to co-operate with the RNN on leaning complex compositions. Additionally, the hidden state of RNN together with external memory in NTM and sequence of embedding vectors are gathered to generate the gated annotation used by the attention mechanism. 

On the other hand, incorporating the embeddings into the attention mechanism can also be viewed as building a shortcut connections that has been proved to alleviate the underlying degradation problem \cite{he2016deep}. Moreover, shortcut connections have an added benefit of not adding any extra parameters or computational complexity.

\subsection{External Memory in NTM}
\label{sec:3_1}
In the conventional RNN based NTM systems, a RNN is used to learn the representation of the sequence. 
Concretely, in the RNN, at each time step $t_i$, the current state $s_i$ is depending on the input $e_i$ from the embedding layer, and the last state $s_{i-1}$. To measure the importance of the input and the historical states, a non-linear function is used to learn the weights of the two parts. In NMT task, most researchers prior to use the GRU benefiting from its simple form. Following the original definition of the GRU, the value of $s_i$ could be calculated as:
 \begin{eqnarray}
s_i &=& \mathrm{GRU}(e_i, s_{i-1}) \\
	&=&(1-z_i) \odot s^{'}_{i} + z_{i} \odot s_{i-1} 
\label{eqn:gru}
\end{eqnarray}
 
 and
 
  \begin{eqnarray}
s^{'}_{i} &=& \mathrm{tanh}(We_{i} + r_{i} \odot (Us_{i-1}))	\\
r_{i} & =& \sigma(W_{r}e_{i}+U_{r}s_{i-1}) \\
z_{i} &=& \sigma(W_{z}e_{i}+U_{z}s_{i-1})
\label{eqn:gru_detail}
\end{eqnarray}

where $W$, $U$, $W_{r}$, $U_{r}$, $W_z$, $U_z$ are trainable parameters.

However, as equation (\ref{eqn:gru}) and (\ref{eqn:gru_detail}) indicates, the current state is depending on the current input embeddings and historical state. In this situation, the RNN has difficulty in well learning both the current lexical semantics and historical dependent relationships. Another drawback is that the RNN knowing nothing about the future information when generating the current state that potentially affects the capturing of long distance dependent relationships.  

\begin{figure}[tp]
\centering
\includegraphics[width=3.2in]{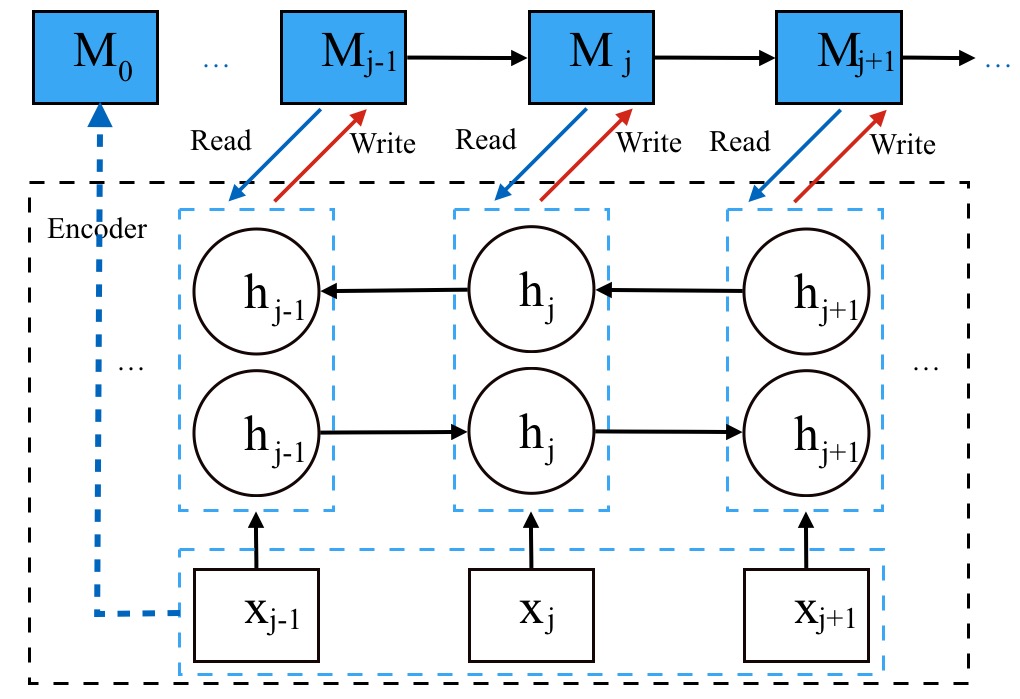}

\caption{Illustration of NTM in the encoder. The RNN reads and writes from the NTM at each time step. }
\label{fig:encoder}
\end{figure}

To enable the encoder to represent the lexical semantics as well as complex composition, our solution is to develop multi-channel encoder that consists of embeddings from original sentence and an external memory in NTM as well as hidden state from the RNN. Most inspired by the design of neural turing machine \cite{graves2014neural} and recent excellent works on external memories for NMT \cite{meng2015deep,wang2016memory,feng2017memory}, we present a \textbf{Read-Write} memory to enhance the GRU in the RNN and intend to capture more complex compositional relationships. 

In order to better understand the whole process, Figure \ref{fig:encoder} illustrates the detailed structure of the encoder. At each time step, the state in RNN queries the external memory using an attention-based addressing mechanism and reads the context of the memory. Instead of directly using the last state to feed the GRU, we use the last state to query the external memory and fetch the contextual memory as the state input of the GRU. This operation guarantees that the controller knowing more contextual information before generating the current state which may potentially assist the decision of GRU. 

Besides the \textbf{Read} operation, we also design a \textbf{Write} operation. The motivation of adding a \textbf{Write} operation is that we expect the RNN and NTM to learn different types of relationships respectively via different updating strategies.    

Formally, let $\mathrm{M}_t \in \mathbb{R}^{n \times m}$ representing the memory in time $t$ after the update of encoding state, where $n$ is the number of memory cells and $m$ is the dimension of vector in each cell. 
We initialize the $\mathrm{M}^0$ by the sequence of embedding vector $\mathrm{E}=[e_1, e_2, ..., e_t]$. 

Inspired by the work of interactive attention \cite{meng2016interactive}, at each time step $t$, we first generate the \textbf{Read} memory $\tilde{\mathrm{M}}_t$ as follows,
\begin{equation}
\tilde{\mathrm{M}}_t = \mathrm{M}_{t-1}(w^{R}_{t} \odot \mathrm{R}_t) 
\label{eq:read_memory}
\end{equation}

where $\mathrm{M}_{t-1}$ is the content of the memory in the last time step, and $w^{\mathrm{R}}_{t} \in \mathbb{R}^n$ specifies the normalized weights assigned to the cells in $\mathrm{M}_{t-1}$. As described in \citeauthor{graves2014neural}, we can use context-based or address-based addressing mechanism to determine $w^{\mathrm{R}}_{t}$. In practice, we found that a conventional attention model work well in our model, thus we compute the $w^{\mathrm{R}}_{t}$ as described in equation (\ref{eq:alpha_ij}) and (\ref{eq:e_ij}). In addition, $\mathrm{R}_t$ is similar to the read gate in the GRU, which determines the content to be read from memory cells. More specifically, it can be defined as,

\begin{equation}
\mathrm{R}_t = \sigma (W_rs_{t-1})
\label{eq:read_gate}
\end{equation}

 where $s_{t-1}$ is the state of the last time step in the encoder, and $W_r  \in \mathbb{R}^{m \times m}$ is trainable parameter.

Once we have obtained the \textbf{Read} memory $\tilde{\mathrm{M}}_t$, we use it to fetch the context $c_t$, similar to the computation specified in equation (\ref{eq:c_i}).
\begin{equation}
c_t = \sum^{T_x}_iw^{R}_{t}\tilde{\mathrm{M}}_t
\label{eq:ntm_c_t}
\end{equation}

After that, $c_t$ is combined with embedding vector $e_t$ and used to update the new state
\begin{equation}
s_t = \mathrm{GRU}(c_t, e_t)
\label{eq:ntm_state}
\end{equation}

Finally, the new state $s_t$ is used to update the external memory by writing to it to finish the round of state-update.
 \begin{equation}
\mathrm{M}^t = \tilde{\mathrm{M}}_t +  w^{W}_{t} \odot \mathrm{U}_t
\label{eq:ntm_write}
\end{equation}

where $U_t = \sigma(W_us_t)$ is the update gate, and is parameterized with $W_u \in \mathbb{R}^{m \times m}$.
Meanwhile, in our experiments, the weights for reading $w^R_t$ and writing $w^W_t$ at time $t$ are shared, that means we compute it using the formulas in equation (\ref{eq:alpha_ij}) and (\ref{eq:e_ij}) with the same state $s_{t-1}$ and the memory $\mathrm{M}_{t-1}$.  

It is worth noting, since we use biRNN to represent the source sequence, we will obtain two external memories: $\overrightarrow{\mathrm{M}}$ and $\overleftarrow{\mathrm{M}}$, which is originated from the forward RNN and the backward RNN, respectively. Similar to the method presented in the traditional NMT, we concatenate the two types of hidden states. We equally concatenate two external memories and use it as one annotation for the attention mechanism. 

\subsection{Gated Annotation}
\label{sec:3_2}
As described in the beginning of the section, we design multiple encoding components in our encoder. Consequently, we will obtain multiple annotations from the encoder, including $\mathrm{M}_t$ from external memory in NTM, $\mathrm{h}_t$ from the hidden state of the RNN and the sequence of embedding vectors $\mathrm{E}=[e_{x_1}, ..., e_{x_t}]$ from the original source input respectively. 

To utilize multiple annotations, one feasible solution is summing or concatenating of them. In this paper, motivated by the design of GRU, we propose an alternative reasonable solution. Since in prior, we can not determine which encoding component is better for the translation, the best decision is to let the model learning the weight between two annotations automatically.

Formally, given the external memory $\mathrm{M} \in \mathbb{R}^{n \times 2 \times m}$ in NTM, and the hidden state $\mathrm{h} \in \mathbb{R}^{n \times 2 \times d}$ in RNN, in which $d$ indicates the hidden size. \footnote{Since we have concatenated the memories and the annotations of the forward and the backward RNN, thus the dimension of two units should be multiplied by two.}
The computation of annotation which will be further utilized by the attention mechanism can be specified as,
\begin{equation}
\mathrm{h}_{rnn\_ntm} = g_0 \odot \mathrm{M} + (1-g_0) \odot \mathrm{h}
\label{eq:rnn_ntm}
\end{equation}

where $g_0$ is the gated unit, calculated as,
\begin{equation}
g_0 = \sigma(W_{g_0}\mathrm{M}+U_{g_0}\mathrm{h})
\label{eq:gate_g}
\end{equation}
where $W_{g_0}$ and $U_{g_0}$ are trainable parameters.

In the experimental section, we will investigate which type of encoding component plays an important role in improving the translation, thus we will validate various of combination of annotations.

Besides the combination of external memory in NTM and hidden state in RNN, we also generate the following combinations:
 \begin{eqnarray}
 \label{eq:rnn_emb}
\mathrm{h}_{rnn\_emb} &=& g_1 \odot \mathrm{E} + (1-g_1) \odot \mathrm{h}	\\
\label{eq:ntm_emb}
\mathrm{h}_{ntm\_emb} &=& g_2 \odot \mathrm{E} + (1- g_2) \odot \mathrm{M}	\\
\label{eq:ntm_rnn_emb}
\mathrm{h}_{ntm\_rnn\_emb} &=& g_3 \odot \mathrm{E} + (1-g_3) \odot \mathrm{h}_{rnn\_ntm} 	
\end{eqnarray}

and
 \begin{eqnarray}
 \label{eq:rnn_emb_gate1}
g_1 &=& \sigma(W_{g_1}\mathrm{E}+U_{g_1}\mathrm{h})	\\
\label{eq:rnn_emb_gate2}
g_2 &=& \sigma(W_{g_2}\mathrm{E}+U_{g_2}\mathrm{M})	\\
\label{eq:rnn_emb_gate3}
g_3 &=& \sigma(W_{g_3}\mathrm{E}+U_{g_3}\mathrm{h}_{rnn\_ntm})		
\end{eqnarray}

where $W_{g_1}$, $U_{g_1}$,$W_{g_2}$, $U_{g_2}$, $W_{g_3}$ and $U_{g_3}$ are trainable parameters.

\section{Experiments}
We mainly evaluate our approaches on the widely used NIST Chinese-English translation task. In order to compare our model to the previous works, we also provide results on the WMT English-French translation task. 
For Chinese-English task, we apply case-insensitive NIST BLEU. For English-French, we tokenize the reference and evaluate the performance with multi-bleu.pl\footnote{\url{https://github.com/moses-smt/\\mosesdecoder/blob/master/scripts/generic/multi-bleu.perl}}. The metrics are exactly the same as in the previous literatures.

\subsection{Data sets}
\noindent 
\textbf{NIST Chinese-English.}
We use a subset of the data available for NIST OpenMT08 task \footnote{1LDC2002E18, LDC2002L27, LDC2002T01,
LDC2003E07, LDC2003E14, LDC2004T07, LDC2005E83,
LDC2005T06, LDC2005T10, LDC2005T34, LDC2006E24,
LDC2006E26, LDC2006E34, LDC2006E86, LDC2006E92,
LDC2006E93, LDC2004T08(HK News, HK Hansards )}. The parallel training corpus contains 1.5 million sentence pairs after we filter with
some simple heuristic rules, such as sentence being too long or containing messy codes. We choose NIST 2006 (NIST06) dataset as our development
set, and the NIST 2003 (NIST03), 2004 (NIST04)
2005 (NIST05), 2008 (NIST08) and 2012 (NIST12)  datasets as our
test sets. We use a source and target vocabulary with 30K most frequent words and filter the sentences longer than 50.

\noindent 
\textbf{WMT'14 English-French.}
We use the full WMT' 14 parallel corpus as our training data. The detailed data sets are Europarl v7, Common Crawl, UN, News Commentary, Gigaword. In total, it includes 36 million sentence pairs. The news-test-2012 and news-test-2013 are concatenated as our development set, and the news-test-2014 is the test set. Our data partition and data preprocess is consistent with previous works on NMT \cite{luong2015effective,jean2014using} to ensure fair comparison. As vocabulary we use 40K sub-word tokens \cite{DBLP:journals/corr/SennrichHB15} based on byte-pair encoding and filter the sentences longer than 120.

\subsection{Model Settings}
For the Chinese-English task, we run widely used open source toolkit DL4MT together with two recently published strong open source toolkits T2T \footnote{\url{https://github.com/tensorflow/tensor2tensor}} and ConvS2S \footnote{\url{https://github.com/facebookresearch/fairseq}} on the same experimental settings to validate the performance of our models\footnote{Parameters for DL4MT: `dim': 1000, `optimizer': `adadelta', `dim\_word': 620, `clip-c': 1.0, `n-words': 30000, `learning-rate': 0.0001, `decay-c': 0.0 \\Parameters for T2T: `model': `transformer', `hparams\_set': `transformer\_base\_single\_gpu' \\Parameters for ConvS2S: `model': `fconv', `nenclayer': 12, `nlayer': 12, `dropout': 0.2, `optim': `nag', `lr': 0.25, `clip': 0.1, `momentum': 0.99, `bptt': 0, `nembed': 512, `noutembed': 512, `nhid': 512}. Beyond that, we also reimplement an attention-based NMT written in tensorflow \footnote{\url{https://www.tensorflow.org/} }as our baseline system. 

To measure the importance of different encoding component on the quality of translation, we validate our approach with different model implements, which includes:
\begin{itemize}
\item DL4MT: an open source toolkit.

\item RNN: an in house implemented attention-based RNN written in tensorflow, and the annotation for attention mechanism consists of only the hidden state of RNN.
\item NTM: using the external memory in NTM directly as the annotation.
\item EMB: using the sequence of embedding vectors as the annotation.
\item NTM-EMB: using the combination of external memory in NTM and embeddings described in the equation (\ref{eq:ntm_emb}).
\item NTM-RNN: using the combination of external memory in NTM and hidden states of RNN described in the equation (\ref{eq:rnn_ntm}).
\item RNN-EMB: using the combination of annotation from hidden state of RNN and embeddings described in the equation (\ref{eq:rnn_emb}).
\item NTM-RNN-EMB: using the combination of overall three encoding components described in the equation (\ref{eq:ntm_rnn_emb}).
\item T2T: an open source toolkit\cite{DBLP:journals/corr/VaswaniSPUJGKP17} .
\item ConvS2S: an open source toolkit\cite{gehring2017convolutional} .
\end{itemize}

In that the main intent of our work is to confirm that using multiple-channel encoder do help on the quality of translation, thus we utilize the empirical hyper parameters according to our previous research.
In more details, we use 512 dimensional word embeddings for both the source and target languages. All hidden layers both in the encoder and the decoder, have 512 memory cells. The output layer size is the same as the hidden size. The dimension of $c_j$ is 1024.

\begin{table*}[htb]

\begin{center}
\begin{tabular}{c|r|c|c|c|c|c|c}
SYSTEMS &PARAMETERS& NIST03 & NIST04 & NIST05  & NIST08 & NIST12 & AVG \\
\hline
DL4MT &- &31.82 &34.86 &31.81 &24.71  &20.47 &28.73 \\

 \hline
RNN&56,314,881 &36.65 &39.86 &36.75 & 29.76& 25.34&33.67(+4.94) \\
NTM&59,988,995  &37.73 &40.03 &36.06 &29.19  &25.12 &33.63(+4.90) \\
EMB&58,416,131  &21.43 &24.38 &20.58 &16.71  &14.48 &19.52 \\
\hline
NTM-EMB&63,135,747  &36.97 &40.05 & 36.35&29.87 &25.51 &33.75(+5.02) \\
NTM-RNN &64,184,323 &37.63 & 40.44&37.77 &30.44 &25.38 &34.33(+5.60) \\
RNN-EMB&59,461,633  &37.98 &40.90 &37.59 &30.58 &25.99 &34.61(+5.88) \\
\hline
NTM-RNN-EMB&67,331,075 &38.56 &40.79 &38.49 &31.51  &26.90 &\textbf{35.25(+6.52)}\\
\hline
T2T  &-&38.11&41.41 &38.12 &31.53  &25.55 &34.94 \\
ConvS2S &-&38.85&40.79 &37.44 &30.60  &26.08 &34.75 \\
 
\end{tabular}
\end{center}

\caption{Performance of different systems on the NIST Chinese-to-English translation tasks. Compared to the strong open source system DL4MT, our models achieve significant improvements. We also list results from another two recently published open source toolkits T2T and ConvS2S for comparison. Remember that both T2T and ConvS2S are deep models with multiple layers.}
\label{tbl:zh-en}
\end{table*}
\subsection{Training Details}
Inspired by the work of GNMT \cite{DBLP:journals/corr/WuSCLNMKCGMKSJL16}, 
we initialize all trainable parameters uniformly between [-0.04, 0.04]. As is common wisdom in training
RNN models, we apply gradient clipping: all gradients are uniformly scaled down such that
the norm of the modified gradients is no larger than a fixed constant, which is 1.0 in our case. 

Following the work of \cite{DBLP:journals/corr/VaswaniSPUJGKP17}, we use the Adam optimizer with $\beta_1=0.9$, $\beta_2=0.98$ and $\epsilon=10^{-9}$.We use the similar learning rate setting with minor modification to adapt our parallel training equipment. 
Over the course of training, the learning rate is calculated according to the formula:
\begin{eqnarray}
lrate &=& d^{-0.5} \cdot \mathrm{min}(\alpha^{-0.5}, \alpha \cdot \beta^{-1.5}) \\
\alpha &=& \mathrm{step\_num} / \mathrm{num\_gpus}
\label{eq:lrate}
\end{eqnarray}
\begin{table*}[htp]

\begin{center}
\begin{tabular}{l|l|c}
SYSTEMS  & Voc.  & En-Fr \\
\hline

 \multicolumn{3}{c}{Existing RNN systems} \\
 \hline
LSTM (6 layers)\cite{luong2015effective} & 80K & 31.50 \\
LSTM (6 layers + PosUNK)\cite{luong2015effective}  & 80K & 33.10 \\
Deep-Att\cite{zhou2016deep} & 80K & 37.70 \\
Deep-Att + PosUnk\cite{zhou2016deep} & 80K& \textbf{39.20} \\
GNMT WPM-32K\cite{DBLP:journals/corr/WuSCLNMKCGMKSJL16} & 80K & 38.95 \\
DeepLAU + PosUNK trained on 12M data\cite{wang2017deep} & 80K &35.10 \\
GNMT WPM-32K, HyperLSTM \cite{DBLP:journals/corr/HaDL16} & 80K & \textbf{40.03} \\
\hline
 \multicolumn{3}{c}{Existing Other systems} \\
 \hline
 ConvS2S (15 layers) + BPE-40K\cite{gehring2017convolutional}  & 40K & 40.46 \\
 Transformer(base)\cite{DBLP:journals/corr/VaswaniSPUJGKP17} & - & 38.10\\
 Transformer(big)\cite{DBLP:journals/corr/VaswaniSPUJGKP17} & - & \textbf{41.00}\\
\hline
 \multicolumn{3}{c}{Our system} \\
 \hline
  RNN + BPE-40K & 40K & 38.19 \\
   MCE + BPE-40K & 40K & 38.80 \\
\end{tabular}
\end{center}

\caption{English-to-French task: BLEU scores. The \textit{RNN} is our basic RNN model, and the \textit{MCE} model combines three encoding components from embeddings, hidden state from RNN, external memory in NTM. Noting that our model does not perform PosUNK and use small size of vocabulary.}
\label{tbl:ef-eg}
\end{table*}

where we set the $\beta=6000$ represents the warm steps with the same meaning in the original work.
Since we train our model with parallelization at the data batch level, we penalize the number of steps by the division of the number of GPUs used in our model.
As we set the batch size to 128, on Chinese-English task it takes around 1 day to train the basic model on 8 NIVDIA P40 GPUs and on English-French task it takes around 7 days.

Additionally, translations are generated by a beam search and log-likelihood scores are normalized by sentence length. And we use a beam width of 10 in all the experiments. Moreover, dropout is also applied on the output layer to avoid over-fitting and we set the dropout rate to 0.5.

\subsection{Results on Chinese-to-English}
Table \ref{tbl:zh-en} lists the overall results on each Chinese-English evaluation tasks. To confirm that our system is strong enough, we also report performance of an open source system: DL4MT. 
At first, we find that there is a significant improvements of our systems over the DL4MT. 
Compared to the DL4MT, our basic \textit{RNN} system achieves an improvement by 4.94 BLEU points. 
Although our \textit{RNN} is a basic attention-based NMT, we assemble it with some advanced techniques, such as initialization all parameters uniformly, adding biases for the embedding vectors, using the output of forward RNN as the input of the backward RNN and training with dynamic learning rate.
Whatever, we give the comparison between our basic \textit{RNN} system and open source toolkits is to prove that our baseline is strong enough, and all improvements over baseline system are reliable. 

Unsurprisingly, the \textit{EMB} model which only using embeddings as the annotation obtains a very low performance due to the encoding of no composition of the sentence. More interesting, any complex encoding component combined with embeddings, such as \textit{NTM-EMB} and \textit{RNN-EMB} receives better performance against the individual one. The reason is although both the external memory in NTM and the RNN encode lexical semantics and complex compositions, however at each time step, the lexical semantics will be blended by the history state, thus it is difficult for them to encode the sequence at varying levels. 
While under the architecture of our MCE, it is possible for the decoder to take the source word directly from the embedding component.
 
We also notice that the performance of \textit{NTM} is almost equal to the \textit{RNN} while the performance of \textit{RNN-EMB} is relatively much better than the \textit{NTM-EMB}. One explanation is that the external memory in NTM is initialized with the embeddings, and at each time step an attention-based content addressing mechanism is employed to update the contents of the memory. That means, compared to the RNN, the external memory records more lexical semantics with no composition. Thus when combined with the embeddings encoding component, the RNN can focus on learning compositional relationships and yielding more improvements. 
 
Lastly, when using three encoding components, our best model achieves 1.58 improvements over the strong baseline system and 6.52 BLEU points over the DL4MT, which proves that our model is effective in practice. Noting that, the performance of our best system is also slightly better than the T2T.

\subsection{Results on English-French}
The results on English-French translation are presented in table \ref{tbl:ef-eg}. We compare our NMT systems with various other systems including Deep RNN model, Deep CNN model and Deep Attention model. For fair comparison, here we just list the single system results reported in their papers. On the English-French translation task, a promising founding is that our system achieves comparable performance over the state-of-the-art systems, even compared with the deep models. Besides, compared to other RNN models, our system is very competitive, although our system is a shallow model. 

Noting that, the best reported result of single system on this dataset is from the work of \cite{DBLP:journals/corr/VaswaniSPUJGKP17}, where they build a big attention neural networks with the depth of 6 layers and 16 heads. Compared to the basic version of them: \textit{Transformer(base)}, the performance of our system is even better despite our model is a shallow networks and their system is more deeper.

Clearly, compared to the previous reported works, our model seems being competitive both on the small and large training data. More important, our model is easy to implement. One thing we could conclude is that when installed with advanced training techniques and innovative model designs, RNN based model is still competitive to other models especially on the large scale training corpus.

\begin{figure}[tbp]
\centering
\includegraphics[width=3.2in]{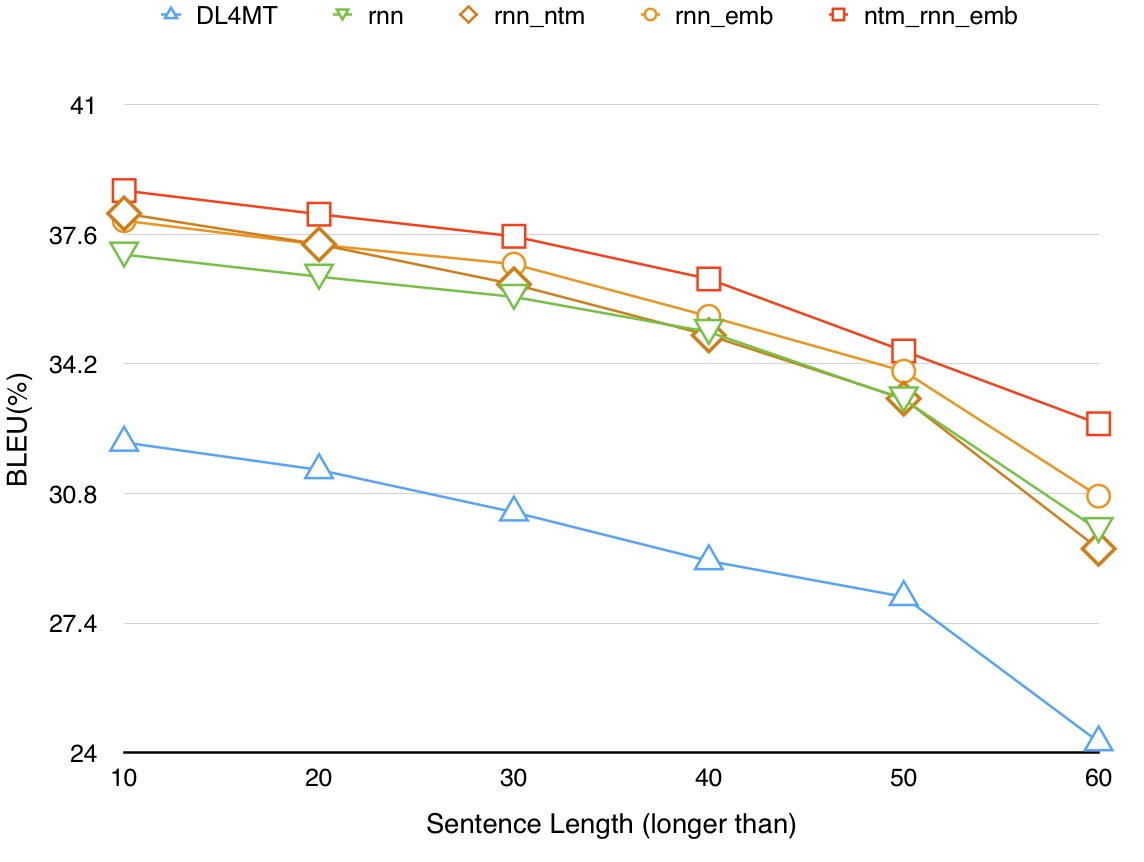}

\caption{Performance on different sentence length.}
\label{fig:length}
\end{figure}

\subsection{Analysis}
On the Chinese-English translation task, we randomly select 1000 sentences from the different testing sets and split the testing set into the different parts according to the length of the sentences.
In particular, we test the BLEU scores on sentences longer than {10; 20; 30; 40; 50; 60} on the test set shown in figure \ref{fig:length}. Clearly, in all curves,
performance degrades with increased sentence length. However, our best model assembled with multiple encoding components yield
consistently higher BLEU scores than the baseline model on longer sentences. These observations are consistent with our intuition
that the MCE can encodes sentence at varying levels which is useful during the translation of long sentence.
 
 \subsection{Translation Sample}
Table  \ref{tbl:sample} shows a particular translation case in which indicates that our model could take pieces of source sentence at varying composition levels.  In this translation case, the model should learn the translation of two named entities `\textbf{Hua Ren Zhi Ye}' and `\textbf{Gao Sheng Ou Zhou Zong Bu}'. Moreover, there is one long distance reordering relationship between `\textbf{Gao Jia}' and  `{\textbf{Mai Xia}'. Thus it is challenging for models to translate this sentence correctly.

From the table \ref{tbl:sample} we find that one-channel models \textit{RNN} and \textit{NTM} failed to translate the `\textbf{Gao Jia}', while the multi-channel models excepts \textit{NTM-RNN} successfully capture the reordering relationship and translate the  `\textbf{Gao Jia}' in correct form. One reason for the poor performance of \textit{NTM-RNN} is that it lacks one component to take the original word embedding for raw encoding with no composition, thus affects the NTM and RNN components to learn the long distance compositional relationship.
\begin{table*}[htp]
   
\begin{center}
\begin{tabular}{r|l}
Source: & {\color{blue} Hua Ren Zhi Ye}  {\color{green} Gao Jia} {\color{red}Mai Xia} Gao Sheng Ou Zhou Zong Bu . \\ 
Reference: & {\color{blue} Chinese Estates}  {\color{red} buys} Goldman’ European Headquarters {\color{green}at high price} . \\ 
\hline
RNN: & {\color{blue}Chinese Home Buyers} {\color{red}purchase} {\color{green}high prices of} OOV european headquarters . \\
NTM: & {\color{blue}Chinese Home Buyers} {\color{red}buy} OOV european headquarters . \\
NTM-RNN: & {\color{blue}Chinese} {\color{red}purchase} {\color{green}prices at high price}  {\color{red}to buy} OOV headquarters in europe .\\
\hline
RNN-EMB: & {\color{blue}China’s Home Buyers} {\color{red}purchased} OOV european headquarters {\color{green}at high price} .\\
NTM-EMB: & {\color{blue}Chinese} {\color{red}bought} OOV european headquarters {\color{green}at high price} .\\
\hline
NTM-RNN-EMB: & {\color{blue}Chinese Home Buyers} {\color{red}purchase} OOV european headquarters {\color{green}at high price} .\\
\end{tabular}
\end{center}
\caption{Translation results for each models. OOV represents the word which is out of vocabulary.}
\label{tbl:sample}
\end{table*}

\section{Conclusion and Future Work}
In this paper, we have proposed multiple-channel encoder to enhance the encoder and the attention mechanism of attention-based neural machine translation. 
To enable the encoder to represent the sentence at varying levels, beside the hidden state of the RNN, we employ the original word embedding for raw encoding with no composition, and design a particular external memory in Neural Turing Machine for more complex composition. A gated annotation mechanism is developed to learn the weights of different encoding components automatically. 
Experiments on extensive Chinese-English translation tasks show that MCE improve the quality of translation and our best model beat the DL4MT by 6.52 BLEU points. And fair comparison on English-French translation indicates that our shallow RNN based model obtain the comparable performance against the previous reported literatures.

In the future, we will attempt to develop more encoding components, such as CNN \cite{lecun1995convolutional} and self-attentive structure\cite{DBLP:journals/corr/LinFSYXZB17} .
\section{Acknowledgements}
This research is supported by the National Basic Research
Program of China (973 program No. 2014CB340505). We
would like to thank Zhengdong Lu, Daqi Zheng and the anonymous reviewers
for their insightful comments.
\bibliographystyle{aaai} 
\bibliography{aaai}

\begin{thebibliography}{}

\bibitem[\protect\citeauthoryear{Cho \bgroup et al\mbox.\egroup
  }{2014}]{cho2014learning}
Cho, K.; Van~Merri{\"e}nboer, B.; Gulcehre, C.; Bahdanau, D.; Bougares, F.;
  Schwenk, H.; and Bengio, Y.
\newblock 2014.
\newblock Learning phrase representations using rnn encoder-decoder for
  statistical machine translation.
\newblock {\em arXiv preprint arXiv:1406.1078}.

\bibitem[\protect\citeauthoryear{Feng \bgroup et al\mbox.\egroup
  }{2017}]{feng2017memory}
Feng, Y.; Zhang, S.; Zhang, A.; Wang, D.; and Abel, A.
\newblock 2017.
\newblock Memory-augmented neural machine translation.
\newblock {\em arXiv preprint arXiv:1708.02005}.

\bibitem[\protect\citeauthoryear{Gehring \bgroup et al\mbox.\egroup
  }{2017}]{gehring2017convolutional}
Gehring, J.; Auli, M.; Grangier, D.; Yarats, D.; and Dauphin, Y.~N.
\newblock 2017.
\newblock Convolutional sequence to sequence learning.
\newblock {\em arXiv preprint arXiv:1705.03122}.

\bibitem[\protect\citeauthoryear{Graves, Wayne, and
  Danihelka}{2014}]{graves2014neural}
Graves, A.; Wayne, G.; and Danihelka, I.
\newblock 2014.
\newblock Neural turing machines.
\newblock {\em arXiv preprint arXiv:1410.5401}.

\bibitem[\protect\citeauthoryear{Ha, Dai, and
  Le}{2017}]{DBLP:journals/corr/HaDL16}
Ha, D.; Dai, A.~M.; and Le, Q.~V.
\newblock 2017.
\newblock Hypernetworks.
\newblock {\em CoRR} abs/1609.09106.

\bibitem[\protect\citeauthoryear{He \bgroup et al\mbox.\egroup
  }{2016}]{he2016deep}
He, K.; Zhang, X.; Ren, S.; and Sun, J.
\newblock 2016.
\newblock Deep residual learning for image recognition.
\newblock In {\em Proceedings of the IEEE conference on computer vision and
  pattern recognition},  770--778.

\bibitem[\protect\citeauthoryear{Hochreiter and
  Schmidhuber}{1997}]{hochreiter1997long}
Hochreiter, S., and Schmidhuber, J.
\newblock 1997.
\newblock Long short-term memory.
\newblock {\em Neural computation} 9(8):1735--1780.

\bibitem[\protect\citeauthoryear{Jean \bgroup et al\mbox.\egroup
  }{2014}]{jean2014using}
Jean, S.; Cho, K.; Memisevic, R.; and Bengio, Y.
\newblock 2014.
\newblock On using very large target vocabulary for neural machine translation.
\newblock {\em arXiv preprint arXiv:1412.2007}.

\bibitem[\protect\citeauthoryear{LeCun, Bengio, and
  others}{1995}]{lecun1995convolutional}
LeCun, Y.; Bengio, Y.; et~al.
\newblock 1995.
\newblock Convolutional networks for images, speech, and time series.
\newblock {\em The handbook of brain theory and neural networks} 3361(10):1995.

\bibitem[\protect\citeauthoryear{Lin \bgroup et al\mbox.\egroup
  }{2017}]{DBLP:journals/corr/LinFSYXZB17}
Lin, Z.; Feng, M.; dos Santos, C.~N.; Yu, M.; Xiang, B.; Zhou, B.; and Bengio,
  Y.
\newblock 2017.
\newblock A structured self-attentive sentence embedding.
\newblock {\em CoRR} abs/1703.03130.

\bibitem[\protect\citeauthoryear{Luong, Pham, and
  Manning}{2015}]{luong2015effective}
Luong, M.-T.; Pham, H.; and Manning, C.~D.
\newblock 2015.
\newblock Effective approaches to attention-based neural machine translation.
\newblock {\em arXiv preprint arXiv:1508.04025}.

\bibitem[\protect\citeauthoryear{Meng \bgroup et al\mbox.\egroup
  }{2015}]{meng2015deep}
Meng, F.; Lu, Z.; Tu, Z.; Li, H.; and Liu, Q.
\newblock 2015.
\newblock A deep memory-based architecture for sequence-to-sequence learning.
\newblock {\em arXiv preprint arXiv:1506.06442}.

\bibitem[\protect\citeauthoryear{Meng \bgroup et al\mbox.\egroup
  }{2016}]{meng2016interactive}
Meng, F.; Lu, Z.; Li, H.; and Liu, Q.
\newblock 2016.
\newblock Interactive attention for neural machine translation.
\newblock {\em arXiv preprint arXiv:1610.05011}.

\bibitem[\protect\citeauthoryear{Schuster and
  Paliwal}{1997}]{schuster1997bidirectional}
Schuster, M., and Paliwal, K.~K.
\newblock 1997.
\newblock Bidirectional recurrent neural networks.
\newblock {\em IEEE Transactions on Signal Processing} 45(11):2673--2681.

\bibitem[\protect\citeauthoryear{Sennrich \bgroup et al\mbox.\egroup
  }{2017}]{uedin-nmt:2017}
Sennrich, R.; Birch, A.; Currey, A.; Germann, U.; Haddow, B.; Heafield, K.;
  {Miceli Barone}, A.~V.; and Williams, P.
\newblock 2017.
\newblock {The University of Edinburgh's Neural MT Systems for WMT17}.
\newblock In {\em {Proceedings of the Second Conference on Machine Translation,
  Volume 2: Shared Task Papers}}.

\bibitem[\protect\citeauthoryear{Sennrich, Haddow, and
  Birch}{2015}]{DBLP:journals/corr/SennrichHB15}
Sennrich, R.; Haddow, B.; and Birch, A.
\newblock 2015.
\newblock Neural machine translation of rare words with subword units.
\newblock {\em CoRR} abs/1508.07909.

\bibitem[\protect\citeauthoryear{Sutskever, Vinyals, and
  Le}{2014}]{sutskever2014sequence}
Sutskever, I.; Vinyals, O.; and Le, Q.~V.
\newblock 2014.
\newblock Sequence to sequence learning with neural networks.
\newblock In {\em Advances in neural information processing systems},
  3104--3112.

\bibitem[\protect\citeauthoryear{Vaswani \bgroup et al\mbox.\egroup
  }{2017}]{DBLP:journals/corr/VaswaniSPUJGKP17}
Vaswani, A.; Shazeer, N.; Parmar, N.; Uszkoreit, J.; Jones, L.; Gomez, A.~N.;
  Kaiser, L.; and Polosukhin, I.
\newblock 2017.
\newblock Attention is all you need.
\newblock {\em CoRR} abs/1706.03762.

\bibitem[\protect\citeauthoryear{Wang \bgroup et al\mbox.\egroup
  }{2016}]{wang2016memory}
Wang, M.; Lu, Z.; Li, H.; and Liu, Q.
\newblock 2016.
\newblock Memory-enhanced decoder for neural machine translation.
\newblock {\em arXiv preprint arXiv:1606.02003}.

\bibitem[\protect\citeauthoryear{Wang \bgroup et al\mbox.\egroup
  }{2017}]{wang2017deep}
Wang, M.; Lu, Z.; Zhou, J.; and Liu, Q.
\newblock 2017.
\newblock Deep neural machine translation with linear associative unit.
\newblock {\em arXiv preprint arXiv:1705.00861}.

\bibitem[\protect\citeauthoryear{Wu \bgroup et al\mbox.\egroup
  }{2016}]{DBLP:journals/corr/WuSCLNMKCGMKSJL16}
Wu, Y.; Schuster, M.; Chen, Z.; Le, Q.~V.; Norouzi, M.; Macherey, W.; Krikun,
  M.; Cao, Y.; Gao, Q.; Macherey, K.; Klingner, J.; Shah, A.; Johnson, M.; Liu,
  X.; Kaiser, L.; Gouws, S.; Kato, Y.; Kudo, T.; Kazawa, H.; Stevens, K.;
  Kurian, G.; Patil, N.; Wang, W.; Young, C.; Smith, J.; Riesa, J.; Rudnick,
  A.; Vinyals, O.; Corrado, G.; Hughes, M.; and Dean, J.
\newblock 2016.
\newblock Google's neural machine translation system: Bridging the gap between
  human and machine translation.
\newblock {\em CoRR} abs/1609.08144.

\bibitem[\protect\citeauthoryear{Zhou \bgroup et al\mbox.\egroup
  }{2016}]{zhou2016deep}
Zhou, J.; Cao, Y.; Wang, X.; Li, P.; and Xu, W.
\newblock 2016.
\newblock Deep recurrent models with fast-forward connections for neural
  machine translation.
\newblock {\em arXiv preprint arXiv:1606.04199}.

\end{thebibliography}
\end{document}